\documentclass{article}
\usepackage{arxiv}
\usepackage[utf8]{inputenc}
\usepackage[T1]{fontenc}
\usepackage{hyperref}
\usepackage{url}
\usepackage{booktabs}
\usepackage{amsfonts}
\usepackage{nicefrac}
\usepackage{microtype}
\usepackage{xcolor}
\usepackage{graphicx}
\usepackage{wrapfig}
\usepackage{amsmath}
\usepackage{cleveref}
\usepackage[table]{xcolor}
\usepackage{booktabs}
\usepackage{listings}
\usepackage{float}
\usepackage{placeins}
\usepackage{caption}
\graphicspath{{figures/}}

\lstdefinestyle{prompt}{
    basicstyle=\ttfamily\scriptsize,
    breaklines=true,
    breakatwhitespace=true,
    columns=fullflexible,
    keepspaces=true,
    showstringspaces=false,
    frame=single,
    framesep=4pt,
    rulecolor=\color{black!30},
    backgroundcolor=\color{black!3},
    xleftmargin=4pt,
    xrightmargin=4pt,
    aboveskip=6pt,
    belowskip=6pt,
}

\lstdefinestyle{toolcode}{
    language=Python,
    basicstyle=\ttfamily\tiny,
    breaklines=true,
    breakatwhitespace=true,
    columns=fullflexible,
    keepspaces=true,
    showstringspaces=false,
    frame=single,
    framesep=3pt,
    rulecolor=\color{black!30},
    backgroundcolor=\color{black!3},
    xleftmargin=3pt,
    xrightmargin=3pt,
    aboveskip=3pt,
    belowskip=3pt,
    keywordstyle=\color{blue!70!black}\bfseries,
    commentstyle=\color{green!40!black}\itshape,
    stringstyle=\color{orange!80!black},
}

\title{ToolFG: Towards Well-Grounded Fine-Grained Image Classification}

\author{
  Yu Xue\textsuperscript{1} \quad Haoxuan Qu\textsuperscript{1}\thanks{Corresponding author} \quad Zhuoling Li\textsuperscript{1} \quad Yihang Lou\textsuperscript{2} \\
  \bf Yan Bai\textsuperscript{2} \quad Hossein Rahmani\textsuperscript{1} \quad Jun Liu\textsuperscript{1} \\
  $^1$Lancaster University \quad $^2$Peking University \\
  {\tt\small \{y.xue9, h.qu5, z.li81, h.rahmani, j.liu81\}@lancaster.ac.uk} \\
  {\tt\small \{yihanglou, yanbai\}@pku.edu.cn}
}

\begin{document}

\maketitle

\begin{abstract}
Fine-grained image classification (FGIC) has broad applications and has attracted significant research attention. In this paper, we explore a novel paradigm for solving FGIC by proposing \textbf{ToolFG}, the first tool-integrated MLLM-based framework tailored to FGIC. ToolFG enables MLLMs to autonomously and flexibly use external tools during the reasoning process, actively interact with images, and collect verifiable visual cues for distinguishing highly similar categories in a more \textit{reliable} and \textit{well-grounded} manner. To equip the model with such tool-use ability, we design a novel \textbf{MCTS-guided tool-use knowledge distillation mechanism}, which effectively mines tool-use- and FGIC-relevant knowledge from advanced proprietary MLLMs for model training. Furthermore, we propose a \textbf{model-tool co-evolution mechanism} that jointly refines the toolset and the model's tool-use policy, driving them toward a mutually adapted and FGIC-specialized state. Extensive experiments demonstrate the effectiveness of our framework.
\end{abstract}

\section{Introduction}
\label{sec:intro}

Fine-grained image classification (FGIC) aims to distinguish highly similar subcategories within the same superordinate category. 
As a fundamental task in computer vision with broad applications such as biodiversity monitoring~\cite{van2015building, van2021benchmarking} and intelligent transportation~\cite{khan2019survey, ke2020fine}, it has attracted sustained research attention~\cite{wei2021fine, he2017fine, he2019fine, he2022transfg, peng2017object, zhang2016picking, yu2018hierarchical}. 
Recently, multimodal large language models (MLLMs) have shown strong visual understanding ability and, through large-scale pretraining, encode rich world knowledge that covers the prototypical attributes defining different subcategories. Motivated by these properties, a growing line of work has applied MLLMs to FGIC~\cite{ kim2024finer, he2025analyzing, hong2025unlabeled, he2026finer, he2026taxonomy, kumar2026divek}. Specifically, these methods design dedicated training recipes that guide MLLMs to attend to subtle discriminative attributes (e.g., fine differences in color, texture, shape, or part proportions), and further elicit chain-of-thought reasoning to analyze the captured visual evidence before arriving at the final classification result~\cite{he2026finer, kumar2026divek}.

However, despite the achieved progress, existing MLLM-based FGIC methods~\cite{he2026finer, kumar2026divek} still remain limited. Specifically, the visual attributes that appear in their reasoning chains often remain vague and qualitative (e.g., ``its tail is not long'').
Such vague attributes are of little help in distinguishing highly similar subcategories, such as the Boat-tailed Grackle and the Rusty Blackbird, whose discrimination usually requires more concrete and accurate measurements (e.g., the relative ratio of tail length to body length).
Without measuring such attributes, the model's reasoning process can be \emph{not well-grounded} and is thus easily misled by its own vague impressions, ultimately resulting in misclassification.
In contrast, human experts are still able to perform FGIC reliably.
We observe that rather than ``only viewing the image'', some human experts will actively interact with the image using image-processing tools (e.g., zooming into key regions, adjusting contrast, and applying sharpening) to better locate and inspect subtle visual attributes, and even further use measurement tools (e.g., color samplers and calipers) to accurately quantify these attributes~\cite{branson2010visual, wah2011multiclass, yang2018learning}, thereby gathering verifiable evidence to ground the final classification decision. For instance, they may observe that the tail is roughly 50\% of the body length, a ratio characteristic of the Boat-tailed Grackle.

\begin{wrapfigure}[17]{r}{0.8\textwidth}
  \centering
  \includegraphics[width=0.8\textwidth]{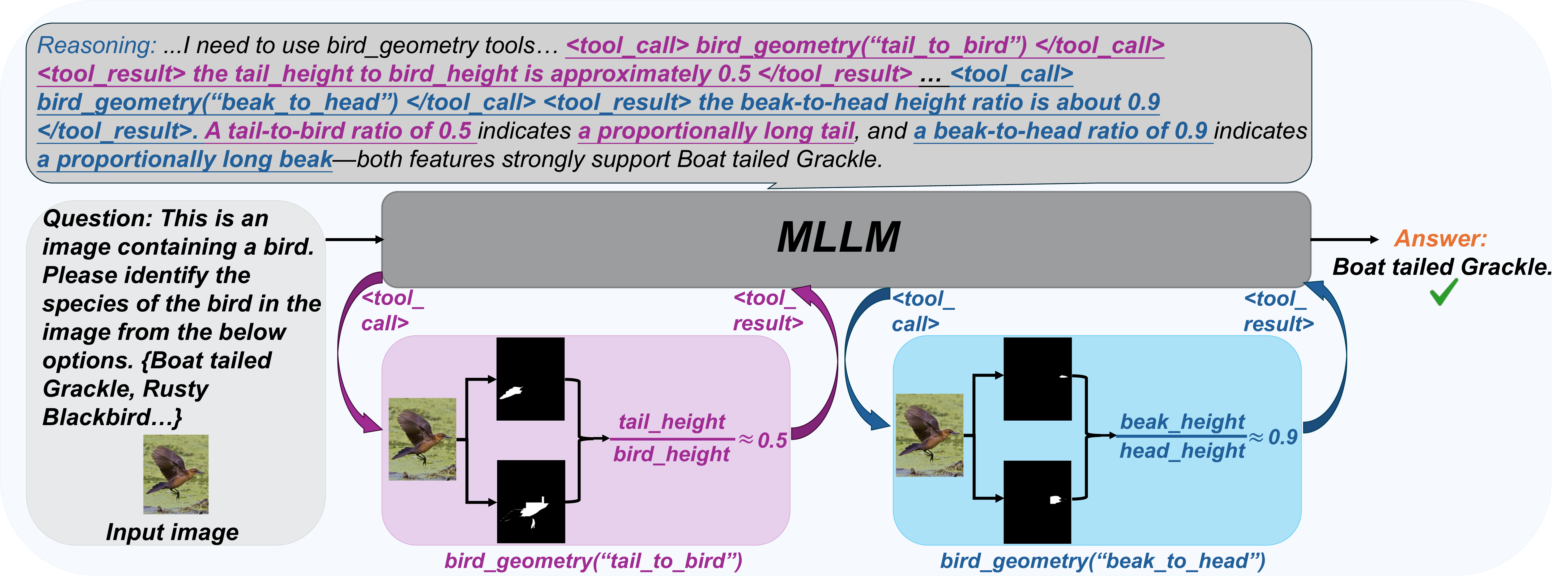}
  \caption{
  Illustration of a tool-integrated reasoning example of our framework, in which the MLLM
  autonomously invokes tools to integrate with the image and collect verifiable evidence before reaching the final FGIC prediction. We show more examples in the supplementary.}
  \label{fig:intro}
\end{wrapfigure}

This naturally motivates us to consider whether MLLM-based FGIC could also benefit from such a human-like tool-using process, which is a very promising direction but still underexplored. Specifically, as shown in~\Cref{fig:intro}, we aim to embed a suite of tools into the MLLM's chain-of-thought reasoning so that at each reasoning step the model can actively interact with images to capture and measure subtle visual attributes, thereby grounding the fine-grained decision (reasoning) process in \textit{verifiable} evidence and ultimately helping the model reliably distinguish highly similar subcategories. However, achieving this is non-trivial. \textbf{First}, it is unclear how to instantiate ``tools'' that allow the MLLM, which has no physical embodiment, to appropriately interact with images and perform verifiable measurements on its own. 
\textbf{More importantly}, even with a well-crafted toolset, teaching the MLLM to proficiently invoke these previously unseen tools for reasoning is itself highly challenging. No supervision is available for tool-integrated reasoning chains, and the only readily accessible signal, the image-level classification label, can be far too sparse to drive direct reinforcement learning over such a high-dimensional sequential decision-making process, in which the model needs to decide when to invoke which tool and in what order.
To tackle the above challenges, we propose \textbf{ToolFG}, a novel tool-integrated MLLM-based framework, which, to the best of our knowledge, is the first method to enable MLLMs to perform FGIC in a \emph{well-grounded} manner by autonomously interacting with images and collecting verifiable visual evidence.
We illustrate the overall training process of our framework in~\Cref{fig:method}, and outline our framework as follows.

Overall, in our framework, the MLLM is optimized to actively leverage a set of tools as a medium for interacting with the image to be classified, capturing and measuring subtle visual attributes so that the reasoning process is grounded in concrete, verifiable evidence. 
Inspired by recent progress in code generation~\cite{jiang2026survey, gulmez2026code},
where LLMs and MLLMs have demonstrated strong potential to understand and invoke functions, we instantiate the tools as function-like APIs that the MLLM can invoke by passing arguments. Once invoked, these tools execute corresponding operations on the image in a sandboxed environment and return textual or visual output to the model, which is then incorporated into the MLLM's subsequent reasoning.
For example, the tool \texttt{bird\_color\_histogram} takes as input the name of a bird part (e.g., breast or wing) and returns the HSV and RGB color histograms of its pixels, providing concrete evidence for color-based attribute comparisons.
Building on this design, we further develop an LLM-based tool-creation pipeline
to automatically construct a toolset tailored for FGIC (mentioned in~\Cref{subsec:init}).
With this toolset in hand, the remaining question is how to teach the MLLM to integrate these previously unseen tools into its reasoning chains, so that it can invoke them appropriately to effectively tackle the challenging FGIC task.

The key obstacle to teaching the MLLM to proficiently use tools is the absence of any direct supervision for tool-integrated reasoning chains. 
To tackle this, we introduce a novel \textit{Monte Carlo Tree Search (MCTS)-guided Tool-Use Knowledge Distillation} mechanism. Our key observation is that advanced proprietary MLLMs (e.g., Gemini and GPT), though not tailored for FGIC nor familiar with our constructed toolset, already encode broad yet coarse knowledge that may cover cues about FGIC and tool use, and can sometimes produce format-valid reasoning chains that correctly invoke our constructed tools, from which they can sometimes even arrive at correct classification results.
Thus, we argue that these advanced proprietary MLLMs can be treated as \emph{broad-but-not-specialized teachers}. Building on this insight, as the core component of our tool-use knowledge distillation mechanism, we design an MCTS-guided exploration process that steers such a teacher MLLM toward promising tool-integrated reasoning chains that lead to correct FGIC predictions, which then serve as supervision for training a \emph{specialized student} (typically a smaller-capacity MLLM) that becomes proficient at tool use for FGIC. In this way, knowledge genuinely relevant to FGIC and tool use can be effectively distilled from the teacher's broad experience and imparted to the student model.

Through the above teacher-student-based training, the student model should learn to imitate the teacher's tool-use pattern (e.g., producing syntactically valid tool calls) and acquire familiarity with the appropriate use of the initial toolset. However, as shown in recent studies on LLM training~\cite{yang2025emperor},
such supervision-based training may amount to mere pattern imitation of the supervision signal, and thus its generalization to diverse FGIC scenarios can still be limited.
In addition, a one-shot constructed toolset is also unlikely to be optimal, which may still be insufficient to handle subtle visual attributes encountered across different FGIC samples. Besides, it would be ideal for the toolset to keep pace with the model's gradually improving tool-use ability, so that more powerful tools could be introduced as the model becomes more capable. 
To this end, we further introduce a \textit{model-tool co-evolution mechanism} that iteratively optimizes both the model's tool-use policy and toolset, so that the MLLM and the toolset co-evolve toward a mutually attuned, FGIC-specialized configuration.

Our main contributions are: 1) We propose \textbf{ToolFG}, the first tool-integrated reasoning framework that enables MLLMs to solve FGIC in a \emph{well-grounded} manner.
2) Through the novel \emph{MCTS-guided tool-use knowledge distillation} and \emph{model-tool co-evolution} mechanisms, our framework can effectively tailor tools and train the MLLM to master them for solving the challenging FGIC task.
3) Extensive experiments on multiple evaluation benchmarks show the efficacy of our method.

\section{Related Work}
\label{sec:related}

\noindent\textbf{Fine-grained Image Classification.} Fine-grained image classification (FGIC) has attracted sustained research attention in the computer vision community~\cite{wei2021fine, he2017fine, he2019fine, he2022transfg, peng2017object, zhang2016picking, yu2018hierarchical, luo2019cross, sun2022sim, chen2019destruction, zhuang2020learning, gao2020channel, rao2021counterfactual, zhang2014part, lin2015bilinear, fu2017look, du2020fine}.
Earlier FGIC methods have explored different model architectures for this task, from CNNs~\cite{zhang2014part, lin2015bilinear, fu2017look} to Transformer-based approaches~\cite{he2022transfg, sun2022sim, rao2021counterfactual}. 
Motivated by the rich world knowledge and strong visual reasoning capabilities of MLLMs, recent methods~\cite{he2025analyzing,he2026finer,kumar2026divek} have increasingly adopted MLLMs for solving the challenging FGIC task. 
For example,
Finedefics~\cite{he2025analyzing} improves FGIC by strengthening the alignment between visual representations and fine-grained category names through attribute-augmented contrastive learning. 
DiVE-k~\cite{kumar2026divek} organizes the top-$k$ predictions on each image as a multiple-choice question and uses reinforcement learning with a verifiable reward to force the model to compare among the most confusable candidate subclasses.

Different from them, our work casts MLLM-based FGIC as a process that needs \emph{well-grounded} reasoning, in which the MLLM actively invokes tools to collect concrete visual evidence, so that every step of its reasoning is backed by verifiable evidence rather than its internal and vague impressions.

\noindent\textbf{Reasoning for MLLMs.} Recently, adapting the general reasoning capability of MLLMs into task-specific reasoning patterns has attracted considerable attention across a range of domains~\cite{zhang2023multimodal, zheng2023ddcot, shao2024visual, mitra2024compositional, huang2026visionr,dong2025tool,ma2025automated,yuan2025video}, including reasoning segmentation~\cite{lai2024lisa}, detection-oriented MLLM reasoning~\cite{pi2023detgpt}, and mathematical and chart reasoning~\cite{zhang2025mavis}. 
Different from these works, in this paper, to address the 
challenging FGIC task, we propose a novel \emph{MCTS-guided knowledge distillation} mechanism to mine tool-integrated reasoning knowledge for training the MLLM to possess tool-use ability, and further design a new \emph{model-tool co-evolution} mechanism to mutually promote the model and the toolset, making them increasingly powerful and compatible with each other.

\section{Method}
\label{sec:method}

\begin{figure}[t]
  \centering
  \includegraphics[width=\textwidth]{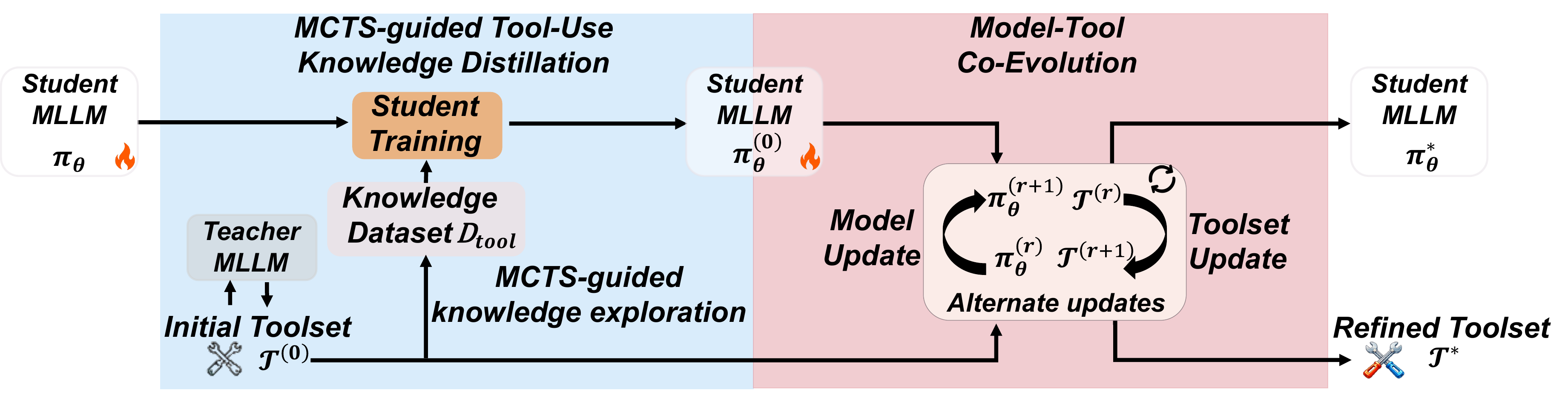}
  \vspace{-0.5cm}
  \caption{Our framework comprises two key mechanisms in its training pipeline that together enable the MLLM to master tool use and acquire FGIC-task-relevant knowledge through two complementary stages. In (1) \emph{MCTS-Guided Tool-Use Knowledge Distillation}, our framework first initializes a toolset $\mathcal{T}^{(0)}$, and then drives an advanced proprietary MLLM teacher to use $\mathcal{T}^{(0)}$ to explore insightful tool-integrated reasoning trajectories, which form a tool-use knowledge dataset $\mathcal{D}_{\text{tool}}$ for training a student policy $\pi_{\theta}^{(0)}$. Besides, in (2) \emph{Model-Tool Co-Evolution}, we further alternately optimize the policy to discover more effective tool-use patterns, and also refine the toolset so that it keeps pace with the improving tool-use capability of the MLLM. After conducting the above two stages, the resulting MLLM policy $\pi_{\theta}^*$ and toolset $\mathcal{T}^{*}$ are ready for performing more well-grounded FGIC.}
\vspace{-0.4cm}
  \label{fig:method}
\end{figure}

\noindent\textbf{Overview.}
Our ToolFG aims to enable MLLMs to perform FGIC in a \emph{well-grounded} manner by invoking tools to interact with the image and collect verifiable evidence.
Yet realizing this paradigm is very challenging.
There is no off-the-shelf well-tailored toolset for FGIC, and, more importantly, even with a ready toolset, teaching the model to autonomously and flexibly invoke these previously unseen tools within a complex reasoning chain remains non-trivial. To tackle this, as shown in~\Cref{fig:method}, our framework introduces two complementary stages in its training pipeline to enable the MLLM to master the tools and also facilitate the construction of an effective toolset.

First, considering no toolset tailored for FGIC is readily available, we design an LLM-based tool-creation pipeline to automatically initialize a set of tools (discussed
in~\Cref{subsec:init}) that, while not necessarily optimal, captures the basic operations a human expert would rely on and serves to bootstrap the MLLM's mastery of the tool-use pattern. With this initial toolset in hand, and given the lack of trajectory-level supervision, we introduce a novel \textbf{ (1) MCTS-Guided Tool-Use Knowledge Distillation} mechanism, in which an advanced proprietary teacher MLLM (e.g.,\ Gemini)
is guided to explore promising tool-integrated reasoning trajectories, which are then used as supervision to train a student MLLM 
specialized in FGIC-tailored tool invocation.
Additionally, we further design an \textbf{(2) Model-Tool Co-Evolution} mechanism to consolidate and promote the generalization of the trained model, and to jointly upgrade the toolset, by refining existing tools and exploring new ones, to keep pace with the model's growing tool-use ability.

Below, we first formalize our proposed tool-integrated FGIC paradigm in~\Cref{sec:formulation}, and then detail the toolset initialization and MCTS-guided tool-use knowledge distillation mechanisms in~\Cref{subsec:init}, followed by the policy--tool co-evolution mechanism in~\Cref{subsec:coevolve}.

\subsection{Formulation of Tool-Integrated FGIC Paradigm}
\label{sec:formulation}

Overall, our goal is to learn an MLLM policy \(\pi_{\theta}\) and equip the MLLM with an appropriate toolset \(\mathcal{T}=\{t_k\}\). Each tool \(t_k\) is associated with a structured tool card \(\mathcal{C}_{t_k}\), which provides a standardized description of the tool's functionality, input arguments, and return values, as detailed in the supplementary. Given a test input image \(x\), our goal is then to derive, using the learned policy and toolset, a tool-integrated reasoning chain \(\tau=\{p,x,a_1,a_2,\dots,a_n\}\) that leads to the correct classification result \(\hat{y}\) for \(x\). Here, \(p\) denotes the system prompt containing the task description and tool cards, while \(a_i\) denotes the action at step \(i\), which is either a tool invocation with specific arguments or a termination action that outputs the final prediction \(\hat{y}\).

\subsection{MCTS-guided Tool-Use Knowledge Distillation}
\label{subsec:init}

As stated in the Introduction, although no direct trajectory-level supervision is available for teaching the model to harness tools, existing advanced proprietary MLLMs (e.g.,\ Gemini) may already encode the FGIC- and tool-use-relevant knowledge we need, and can produce reasoning chains that yield correct classification results albeit at a ``relatively low hit rate''. This naturally motivates us to treat such models as \emph{broad-but-not-specialized teachers} and distill the FGIC-specialized tool-use knowledge (i.e.,\ tool-integrated reasoning supervision) from them into a student policy.

With this insight, we introduce a novel MCTS-guided trajectory exploration mechanism, which treats the step-by-step reasoning chain as a search tree and guides the teacher toward more promising trajectories that yield correct results, within the vast space the possible trajectories span. Notably, beyond revealing successful trajectories, the search tree built during exploration also surfaces their sibling paths that appear reasonable at intermediate steps yet ultimately end in wrong predictions. We argue that these failed sibling trajectories should also be collected as \emph{hard negatives} paired with their positive counterparts. Such paired supervision allows the student policy to learn not only to imitate successful tool use, but also to avoid seemingly reasonable yet ultimately flawed tool-use decisions.

In what follows, we instantiate the above idea in three steps. \textbf{Step 1}, we design an LLM-based tool-creation pipeline that automatically constructs an initial toolset $\mathcal{T}^{(0)}$, serving as the action space for all subsequent steps; \textbf{Step 2}, we collect paired positive--negative supervision trajectories over $\mathcal{T}^{(0)}$ from the teacher via a proposed MCTS-guided exploration scheme; and \textbf{Step~3}, we train the student policy for tool invocation through a carefully designed trajectory-level contrastive objective.

\noindent\textbf{Step~1: Automatically Constructing the Initial Tool Set.}
We design an MLLM-based automatic tool construction pipeline. To build the toolset, our pipeline first identifies a set of attributes that are useful for distinguishing highly similar subcategories, and then constructs invocable tools based on these attributes and task-specific knowledge from existing FGIC practices. More details about tool construction are provided in the supplementary.
Through this procedure, we obtain an initial toolset $\mathcal{T}^{(0)} = \{t_k\}$, which serves as the action space for the subsequent trajectory exploration and policy training stages.

\noindent\textbf{Step~2: Exploring Paired Trajectory Supervision.}
In this step, we aim to mine reasoning-trajectory-level supervision to teach the model to harness $\mathcal{T}^{(0)}$ for solving FGIC. 
Inspired by tree search methods~\cite{coulom2006efficient,li2026automatic,yang2026tooltree}, we model the teacher's reasoning process over the toolset $\mathcal{T}^{(0)}$ as a search tree, in which each node represents a partial reasoning chain $\{p, x, a_1, \dots, a_l\}$ and each edge corresponds to a possible next action $a_{l+1}$. We iteratively grow the search tree to expose branches that lead to correct classifications, which serve as positive trajectories $\tau^{+}$, and branches that look reasonable at intermediate steps yet ultimately end in wrong predictions, which serve as hard-negative trajectories $\tau^{-}$. 
At each iteration, we first \emph{select} a promising frontier node from the current tree to decide where to keep searching, then \emph{expand} this node by proposing a small set of candidate next actions to attach as children, \emph{simulate} a complete rollout from each new child to score the long-term impact of taking that action rather than its local appearance, and finally \emph{backpropagate} the rollout score along the visited path to refine the value estimates that guide subsequent selections. This four-operation cycle is conducted iteratively, after which $\tau^{+}$ and $\tau^{-}$ are jointly extracted from the search tree. We detail each step below.

\quad\underline{Selection.}
After the search has run for a few iterations, the search tree will contain many candidate branches, and each subsequent rollout needs to decide where to continue from. Yet, a naively greedy descent would collapse the search onto a single best path and prune away plausible-but-failing alternatives,  which are exactly the hard negatives we want to mine. We therefore select a \textit{frontier node} (a partially-explored node whose reasoning chain has not yet produced a final classification) by the UCB rule~\cite{kocsis2006bandit}, which balances exploiting high-value branches and exploring less-visited ones in a principled way:
$
a^{*} = \arg\max_a \big[Q(s,a) + c\sqrt{\ln N(s)/N(s,a)}\big],
$
where $Q(s,a)$ estimates the value of choosing action $a$ at state $s$, $N(s,a)$ is its visit count, $N(s)$ is the visit count of state $s$, and $c>0$ controls the exploration strength. Starting from the root, we repeatedly apply this rule until reaching a frontier node $s_l$.

\quad\underline{Expansion.}
Given the selected frontier node $s_l$, we expand the tree by adding a small set of candidate next tool invocations. We ask the teacher MLLM to propose $J$ plausible candidates $\{a_{l+1}^{(j)}\}_{j=1}^{J}$ conditioned on the current interaction history. Each candidate is executed to obtain $r_{l+1}^{(j)}$ and instantiated as a new child node, serving as the starting point of subsequent simulation.

\quad\underline{Simulation.}
To evaluate the potential of a newly expanded child nodes, its incomplete reasoning chain alone is insufficient. We therefore continue the rollout from $s_l$ using the teacher MLLM, generating subsequent reasoning steps until the model emits a final prediction $\hat{y}$, thereby obtaining a complete trajectory $\tau$. The child node can then be scored based on both the classification outcome and the overall quality of the completed reasoning trajectory:
$
R(\tau) = R_{\text{cls}} + \lambda R_{\text{traj}}(\tau)$
where $R_{\text{cls}} = \mathbb{1}[\hat{y} = y]$. The trajectory-quality term $R_{\text{traj}}$ employs an LLM-based judge to evaluate the soundness of the entire reasoning process, penalizing "lucky-guess" trajectories that arrive at the correct answer despite containing obviously flawed intermediate steps. Further details are deferred to the supplementary.

\quad\underline{Backpropagation.} To let the rollout outcomes obtained during simulation shape the search behavior of subsequent iterations, we propagate $R(\tau)$ backward through the visited nodes and refresh their visit counts and $Q$-values accordingly (full update rule deferred to the supplementary). These refreshed statistics are the quantities consumed by the UCB score in the next Selection step.

Iterating the four steps above yields a search tree from which we read out a positive trajectory $\tau^{+}$ (the most-visited root-to-leaf path, corresponding to the highest-value reasoning chain), together with a hard negative $\tau^{-}$ (a sufficiently explored path that nonetheless terminates in an incorrect classification). Because both are drawn from the same tree, they share a long common prefix and diverge only at a few critical decision nodes, yielding precisely the contrastive structure Step~3 will exploit. More trajectory extraction details are deferred to the supplementary.
Running the above trajectory extraction on training samples produces our tool-use knowledge dataset $\mathcal{D}_{\text{tool}} = \{(x_i, \tau_i^{+}, \tau_i^{-})\}_{i=1}^{N}$, where each trajectory $\tau_i^{+/-} = (a_1, r_1, \dots, a_{L_i}, r_{L_i}, \hat{y}_i)$ records a complete tool-invocation sequences. In this way, we \emph{upgrade} the supervision available for each sample: the original binary class label is replaced by a process-level positive/negative pair, supplying the contrastive signal that Step~3 consumes for training.

\noindent\textbf{Step 3: Training the Model on the Tool-Use Knowledge Dataset.}
Equipped with the paired tool-use knowledge dataset $\mathcal{D}_{\text{tool}} = \{(x_i, \tau_i^{+}, \tau_i^{-})\}_{i=1}^{N}$, the remaining question reduces to how to exploit it properly for training the student policy. A naive choice is the token-level cross-entropy loss commonly adopted in MLLM supervised training:
$
\mathcal{L}_{\text{CE}}(\theta) = -\frac{1}{|\mathcal{D}_{\text{tool}}|} \sum_{(x, \tau^{+})} \sum_{t} m_t \log p_\theta(\tau^{+}_t \mid \tau^{+}_{<t}, x),
$
where the mask $m_t = 1$ for model-generated tokens and $m_t = 0$ for tool returns and system instructions. 
However, this objective only teaches the model to ``go toward $\tau^{+}$'' and discards the ``do not go toward $\tau^{-}$'' signal that Step~2 carefully mined. In FGIC, $\tau^{+}$ and $\tau^{-}$ often differ at only a single step's tool choice or argument, which is in fact one of the design goals of our search strategy, so this contrast carries extremely dense discriminative information; throwing it away would waste the main value of Step~2.
To use both signals jointly, we draw on contrastive learning in areas such as person re-identification (ReID)~\cite{hermans2017defense}, where the ``anchor + positive + hard-negative'' formulation is widely validated as one of the most effective ways to exploit stratified contrast. Transferring this idea to the trajectory level, we take the input $x$ as the anchor, $\tau^{+}$ as the positive, and $\tau^{-}$ as the hard negative, and encourage the model to assign a higher likelihood to $\tau^{+}$ than to $\tau^{-}$:
\begin{equation}
\mathcal{L}_{\text{trip}}(\theta) = \frac{1}{|\mathcal{D}_{\text{tool}}|} \sum_{(x, \tau^{+}, \tau^{-})} \max\!\Big(0,\; \gamma - \big[\log p_\theta(\tau^{+} \mid x) - \log p_\theta(\tau^{-} \mid x)\big]\Big),
\end{equation}
where $\gamma > 0$ is a margin, and $\log p_\theta(\tau \mid x)$ is computed with the same mask $m_t$ as in $\mathcal{L}_{\text{CE}}$, so only assistant-generated tokens contribute. Combining the two gives the final objective,
\begin{equation}
\mathcal{L}_{\text{init}}(\theta) = \mathcal{L}_{\text{CE}}(\theta) + \beta \cdot \mathcal{L}_{\text{trip}}(\theta),
\label{eq:init}
\end{equation}
where $\beta > 0$ balances the two terms: $\mathcal{L}_{\text{CE}}$ teaches the model what to do right at the token level, while $\mathcal{L}_{\text{trip}}$ teaches it how to avoid going wrong at the trajectory level. Further training details (loss masking, lightweight paraphrasing of tool descriptions, etc.) are deferred to the supplementary.

Through the high-quality construction of tool-integrated reasoning trajectories (Step~2) and their proper use in contrastive training (Step~3), the trained model (denoted as $\pi_{\theta}^{(0)}$) is expected to have acquired a basic tool-use ability: given a fine-grained image, it can autonomously invoke appropriate tools and collect fine-grained visual evidence for solving the FGIC task.

\subsection{Model-Tool Co-Evolution}
\label{subsec:coevolve}
After the above MCTS-guided tool-use knowledge distillation, we obtain an initial toolset and a student policy  that has acquired preliminary competence in tool usage and FGIC reasoning. We now aim to further evolve both the policy and the toolset: the policy should move beyond rigidly mimicking the teacher's tool-use behavior and continually discover more effective tool-use patterns, while the toolset, one-shot initialized in the previous stage, may still be suboptimal and insufficient to handle the subtle visual attributes across FGIC samples, and should also be upgraded to keep pace with the model's improving tool-use ability.  Ideally, the two should converge to a mutually compatible and effective configuration tailored to the FGIC task.
Jointly optimizing the policy and the toolset, however, is intractable for two reasons. \textit{First}, the two are mutually dependent: refining the policy presupposes a suitable toolset, while the suitability of a toolset can only be judged through actual usage by the policy. \textit{Second}, they live in incompatible optimization regimes—the policy is a continuous, differentiable network amenable to gradient-based methods, whereas the toolset is a discrete collection of executable programs that admits no such gradient-based update and can usually be updated through structural operations such as editing or replacement.

To this end, we thus propose a novel iterative co-evolution scheme that alternates between two update (evolution) stages: \textit{Model Update}, where we refine the model policy while keeping the toolset fixed, and \textit{Toolset Update}, where we evolve the toolset based on the usage feedback collected during model update. Through this alternating process, the policy and the toolset progressively improve and become compatible with each other. Below, we describe the two stages in detail.

\noindent\textbf{Model Update: Groundedness Reward-guided Reinforcement Learning.}
In this stage, the toolset is fixed, and we focus on refining the policy alone. Considering the policy obtained from tool-use knowledge distillation may still be limited to merely mimicking the teacher's tool-use patterns~\cite{chen2025sft}, we introduce reinforcement learning to further drive it to discover more effective tool-use strategies through repeated interaction with the tool-integrated FGIC environment. The central challenge is to design a principled reward function that faithfully reflects the quality of the policy's tool-use behavior. In our framework, tool-integrated reasoning is a multi-step decision process: the model needs to decide which tool to invoke and how to set its arguments. A reward defined only by final classification correctness is therefore too coarse: it struggles to identify which intermediate tool decisions caused success or failure, and may even reward trajectories that arrive at correct answers \emph{by luck} after a flawed process (e.g.,\ invoking a wrong tool or misreading a returned value).

To this end, we design a \emph{groundedness reward} $R_{\text{ground}}$, which evaluates the quality of intermediate tool-use decisions along the trajectory. Together with the classification reward $R_{\text{cls}}$ (the binary correctness of the final classification), it forms a composite reward:
$R(\tau) \;=\; R_{\text{cls}}(\tau) \;+\; \eta \cdot R_{\text{ground}}(\tau),$
where $\eta$ balances the two terms. While $R_{\text{cls}}$ keeps the policy aligned with the ultimate task goal, $R_{\text{ground}}$ assigns a lower return to ``right-by-luck'' trajectories whose answers happen to be correct despite flawed tool-use steps, and a higher return to ``right-by-evidence'' trajectories whose reasoning is sound throughout. Optimizing $R$ therefore drives the policy toward genuinely well-grounded tool use rather than accidental correctness. We defer the detailed computation of $R_{\text{ground}}$ to the supplementary.

Through this RL-based model update stage, the model becomes more proficient at using tools, yet may still be bounded by the sub-optimal toolset itself. We address this next by evolving the toolset.

\noindent\textbf{Toolset Update: Usage-Feedback-Driven Evolution.}
We now turn to updating the toolset, which is non-trivial in two ways: tools are executable programs rather than differentiable model parameters, and they come with no ground-truth labels of what an ``ideal'' update should look like.
To tackle this, we leverage a key observation borrowed from real-world tool design: \emph{a tool's effectiveness can usually be judged by its actual users' experience with it}. In our setting, the policy that just finished the policy-update stage is precisely such a user, and its usage feedback on $\mathcal{T}$ consists of two key signals: which tools it chose to invoke, and whether those invocations led to correct predictions. Together, these capture each tool's actual effectiveness, informing which tools to keep, fix, or replace.
Building on this, we further design a novel LLM-driven tool-evolution mechanism that exploits this signal to update the toolset in a non-differentiable way.
Below, we first describe how to collect the user feedback for each tool, and then detail the LLM-driven tool-evolution mechanism.

\quad\underline{Collecting Usage Feedback.}
For each tool $t_k$, we record the policy's behavior during the policy-update stage with two statistics: the invocation frequency $f(t_k)$ (the fraction of training samples on which $t_k$ is invoked), and the conditional accuracy $a(t_k)$ (the fraction of $t_k$-invoking reasoning chains that lead to correct classifications). These two statistics respectively reflect the policy's preference for the tool and the tool's actual effectiveness when used. Combining them, we partition the tools into four categories: \textbf{(A)} high $f$, high $a$ (proven core tools), \textbf{(B)} high $f$, low $a$ (frequently tried but flawed), \textbf{(C)} low $f$, high $a$ (reliable but overlooked), and \textbf{(D)} low $f$, low $a$ (neither used nor effective). For each category, we perform a targeted update operation, as detailed below.

\quad\underline{LLM-driven Tool Evolution.}
With the usage feedback in hand, an LLM (e.g.,\ Gemini) is prompted to launch a targeted operation on each tool. \emph{The key insight is that, by reading and rewriting each tool's source code, documentation, and usage records, the LLM serves as a gradient-free optimizer for the executable-program toolset}.
Specifically: tools in~(A) are retained; tools in~(D) are discarded; tools in~(B), which are tried but flawed, are repaired (e.g.,\ narrowing the parameter space, handling edge cases); and tools in~(C), which are reliable but overlooked, are revised to improve discoverability (e.g.,\ clarifying their documentation).
Besides refining existing tools, the LLM also brainstorms new ones, such as combining complementary tools---e.g.,\ fusing a ``breast color proportion'' tool and a ``wing morphology'' tool into one analyzing ``breast--wing color contrast patterns''. All revised or newly generated tools need to pass \emph{unit tests}---automated checks that verify expected input--output behavior on representative cases---before being admitted. More details about prompt design and tool evolution operations are in the supplementary.

\begin{table}[t]
    \centering
    \setlength{\tabcolsep}{3.4pt}
    \caption{Quantitative comparison with existing methods on zero-shot base-to-novel generalization (per-dataset training). B $\rightarrow$ Base, N $\rightarrow$ Novel, H $\rightarrow$ Harmonic Mean.}
    \label{tab:zero_shot}
\resizebox{0.9\textwidth}{!}{%
    \begin{tabular}{l|*{15}{c}|*{3}{c}}
    \hline
    & \multicolumn{3}{c}{Flowers} & \multicolumn{3}{c}{CUB} & \multicolumn{3}{c}{Pets} & \multicolumn{3}{c}{Cars} & \multicolumn{3}{c}{Aircraft} & \multicolumn{3}{|c}{Average} \\
    Method & B & N & H & B & N & H & B & N & H & B & N & H & B & N & H & B & N & H \\
    \hline
    Qwen2.5-VL-7B~\cite{Qwen2.5-VL} & 84.2 & 83.8 & 84.0 & 63.3 & 48.2 & 54.7 & 87.5 & 93.3 & 90.3 & 58.9 & 72.9 & 65.1 & 50.3 & 54.3 & 52.3 & 68.9 & 70.5 & 69.7 \\
    MaPLe~\cite{khattak2023maple} & 95.9 & 72.5 & 82.6 & 80.0 & 48.8 & 60.6 & 95.4 & 97.8 & 96.6 & 72.9 & 74.0 & 73.5 & 37.4 & 35.6 & 36.5 & 76.3 & 65.7 & 70.6 \\
    PromptSRC~\cite{khattak2023self} & 98.1 & 76.5 & 86.0 & 84.3 & 49.5 & 62.4 & 95.3 & 97.3 & 96.3 & 78.3 & 75.0 & 76.6 & 42.7 & 37.9 & 40.2 & 79.7 & 67.2 & 73.0 \\
    ViRFT~\cite{liu2025visual} & 89.5 & 87.9 & 88.7 & 70.0 & 60.0 & 64.8 & 92.6 & 92.8 & 92.7 & 64.8 & 76.9 & 70.4 & 66.4 & 67.8 & 67.1 & 76.8 & 77.1 & 76.9 \\
    DiVE-k~\cite{kumar2026divek} & 97.4 & 88.9 & 92.9 & 80.5 & 65.5 & 72.2 & 89.1 & 94.2 & 91.6 & 69.0 & 76.2 & 72.4 & 68.1 & 69.1 & 68.6 & 80.8 & 78.8 & 79.8 \\
    Fine-R1~\cite{he2026finer} & 96.6 & 88.7 & 92.4 & 75.8 & 57.0 & 65.1 & 90.9 & 93.0 & 92.0 & 63.6 & 75.1 & 68.9 & 66.8 & 69.7 & 68.2 & 78.7 & 76.7 & 77.7 \\
    \hline
    Ours & 99.3 & 93.1 & 96.1 & 84.8 & 73.0 & 78.5 & 96.8 & 99.7 & 98.2 & 90.0 & 90.5 & 90.2 & 75.6 & 76.5 & 76.1 & \textbf{89.3} & \textbf{86.6} & \textbf{87.9} \\
    \hline
    \end{tabular}}
    \end{table}

\begin{table}[t]
    \centering
    \setlength{\tabcolsep}{3.4pt}
    \caption{Quantitative comparison with existing methods on zero-shot base-to-novel generalization (mixed-dataset training). B $\rightarrow$ Base, N $\rightarrow$ Novel, H $\rightarrow$ Harmonic Mean.}
    \label{tab:zero_shot_mixed}
\resizebox{0.9\textwidth}{!}{%
    \begin{tabular}{l|*{15}{c}|*{3}{c}}
    \hline
    & \multicolumn{3}{c}{Flowers} & \multicolumn{3}{c}{CUB} & \multicolumn{3}{c}{Pets} & \multicolumn{3}{c}{Cars} & \multicolumn{3}{c}{Aircraft} & \multicolumn{3}{|c}{Average} \\
    Method & B & N & H & B & N & H & B & N & H & B & N & H & B & N & H & B & N & H \\
    \hline
    Qwen2.5-VL-7B~\cite{Qwen2.5-VL} & 84.2 & 83.8 & 84.0 & 63.3 & 48.2 & 54.7 & 87.5 & 93.3 & 90.3 & 58.9 & 72.9 & 65.1 & 50.3 & 54.3 & 52.3 & 68.9 & 70.5 & 69.7 \\
    MaPLe~\cite{khattak2023maple} & 93.8 & 78.2 & 85.3 & 77.8 & 45.8 & 57.7 & 94.2 & 97.8 & 96.0 & 72.7 & 72.4 & 72.5 & 39.0 & 29.8 & 33.8 & 75.5 & 64.8 & 69.7 \\
    PromptSRC~\cite{khattak2023self} & 96.3 & 80.7 & 87.8 & 83.5 & 50.7 & 63.1 & 94.7 & 96.4 & 95.5 & 75.2 & 74.7 & 74.9 & 44.9 & 34.8 & 39.2 & 78.9 & 67.5 & 72.7 \\
    ViRFT~\cite{liu2025visual} & 90.2 & 87.9 & 89.0 & 71.3 & 58.3 & 64.2 & 90.2 & 91.6 & 90.9 & 63.7 & 76.6 & 69.6 & 66.9 & 66.4 & 66.7 & 76.5 & 76.2 & 76.1 \\
    DiVE-k~\cite{kumar2026divek} & 97.4 & 89.9 & 93.5 & 76.8 & 61.3 & 68.2 & 87.8 & 94.7 & 91.1 & 68.5 & 78.5 & 73.1 & 65.5 & 69.7 & 67.5 & 79.2 & 78.8 & 78.7 \\
    Fine-R1~\cite{he2026finer} & 95.4 & 87.5 & 91.3 & 73.2 & 55.3 & 63.0 & 87.8 & 94.2 & 90.9 & 64.0 & 76.0 & 69.5 & 61.2 & 67.9 & 64.4 & 76.3 & 76.2 & 76.2 \\
    \hline
    Ours & 99.3 & 92.6 & 95.8 & 83.0 & 72.7 & 77.5 & 97.6 & 98.6 & 98.1 & 92.7 & 96.8 & 94.7 & 74.4 & 75.9 & 75.1 & \textbf{89.4} & \textbf{87.3} & \textbf{88.3} \\
    \hline
    \end{tabular}}
    \end{table}

After the above operations, we obtain an updated toolset, which serves as the fixed environment for the next policy-update round, launching a new ``tool $\rightarrow$ policy $\rightarrow$ tool'' co-evolution cycle.

\subsection{Overall Training and Inference}
\label{subsec:overall}

\noindent\textbf{Training.}
The training process of our framework consists of two stages: 
(1) training the model to acquire basic tool-use knowledge for solving FGIC; and 
(2) further optimizing the model and toolset.
In the first stage, a toolset $\mathcal{T}$ is initialized via our LLM-based tool-creation pipeline (detailed in Step~1 of Sec.~\ref{subsec:init}). Since no direct reasoning supervision, for each training sample, our framework drives a proprietary MLLM to use tools from $\mathcal{T}$ to try to solve the sample, producing paired reasoning trajectory supervision.
By conducting this process over the entire training set, we obtain a trajectory-level dataset $\mathcal{D}_{\text{tool}}$, which is used to supervise model training with the loss function defined in Eq.~\eqref{eq:init}.
In the second stage, we alternately optimize the model and the toolset. With the toolset fixed, we optimize the model using the composite reward defined in the Model Update step of Sec.~\ref{subsec:coevolve}, followed by the tool evolution process detailed in the Toolset Update step of the same section. After this alternating optimization process, we obtain the final model $\pi_{\theta}^{*}$ and toolset $\mathcal{T}^{*}$ for testing.

\noindent\textbf{Testing.}
Given an input image $x$ and a prompt describing the refined toolset $\mathcal{T}^{*}$, the refined MLLM policy $\pi_{\theta}^{*}$ generates a tool-integrated reasoning chain for FGIC prediction. During reasoning, the model invokes function-based tools by outputting their names together with corresponding arguments. Each tool invocation is executed in a sandbox environment, and the execution result is returned to the MLLM in either textual or visual form, depending on the tool. The model then continues reasoning based on the returned evidence until it produces the final classification prediction.

\section{Experiments}
\label{sec:experiments}

\noindent\textbf{Datasets.}
Following DiVE-k~\cite{kumar2026divek}, we evaluate our method on five widely used fine-grained image classification benchmarks: CUB-200~\cite{wah2011caltech}, Oxford Flowers-102~\cite{nilsback2008automated}, Stanford Cars-196~\cite{krause20133d}, Oxford Pets-37~\cite{parkhi2012cats}, and FGVC Aircraft-100~\cite{maji2013fine}. To construct the base-novel partition, we follow prior works~\cite{khattak2023self}, where the category set of each dataset is split in half, with one half designated as base classes and the other as novel classes. Importantly, the names of the novel classes are never seen during training. Additional details are provided in the supplementary.

\noindent\textbf{Evaluation settings and metrics.}
Following DiVE-k~\cite{kumar2026divek}, we evaluate our method under two zero-shot settings and one few-shot setting. The two \emph{zero-shot} settings are as follows: (1) we train a separate model on the base classes of each of the five datasets independently; (2) we merge the base classes of all five datasets into a single mixed dataset and train a single model on it, in order to assess cross-domain generalization. For the \emph{few-shot} setting, the model is trained with $4$ shots per class. As evaluation metrics, we report the classification accuracy on the base classes and on the novel classes, together with their harmonic mean (HM). Additional details are in the supplementary.

\noindent\textbf{Implementation details.}
For the construction of the tool-invocation dataset, we use the advanced proprietary MLLM Gemini~2.5~Flash-Lite~\cite{comanici2025gemini} as the teacher model. Our student model is Qwen2.5-VL-7B~\cite{Qwen2.5-VL}. In the SFT stage, the per-dataset training step counts under the two zero-shot settings are 2000, the training step count on the mixed dataset is 2400; and the training step count under the few-shot setting is 200. All experiments are conducted on NVIDIA H200 GPUs. Additional implementation details are in the supplementary.

\begin{table}[t]
\centering
\setlength{\tabcolsep}{3.4pt}
\caption{Quantitative comparison of our proposed method under 4-shot setting.}
\label{tab:few_shot}
\resizebox{0.7\textwidth}{!}{%
\begin{tabular}{lcccccc}
\toprule
Model & Oxford Flowers & CUB & Oxford Pets & Stanford Cars & FGVC Aircraft & Average \\
\midrule
Qwen2.5-VL-7B~\cite{Qwen2.5-VL} & 78.4 & 51.6 & 79.1 & 57.9 & 52.5 & 63.9 \\
\midrule
MaPLe~\cite{khattak2023maple} & 84.6 & 60.9 & 95.3 & 68.9 & 32.5 & 68.4 \\
PromptSRC~\cite{khattak2023self} & 92.4 & 70.3 & 94.6 & 76.8 & 36.8 & 74.2 \\
ViRFT~\cite{liu2025visual} & 84.8 & 60.0 & 83.1 & 64.8 & 65.4 & 71.6 \\
DiVE-k~\cite{kumar2026divek} & 88.7 & 63.9 & 85.1 & 66.9 & 69.1 & 74.8 \\
Fine-R1~\cite{he2026finer} & 86.0 & 61.9 & 81.1 & 64.1 & 65.1 & 71.6 \\
\midrule
Ours & 95.8 & 72.0 & 97.3 & 83.4 & 76.6 & \textbf{85.0} \\
\bottomrule
\end{tabular}}
\end{table}

\subsection{Main Results}
\label{subsec:main_results}

In our experiments, we compare our method against two categories of baselines. The first category is the open-source model Qwen2.5-VL-7B~\cite{Qwen2.5-VL}. The second category consists of state-of-the-art FGIC methods. Among them, MaPLe~\cite{khattak2023maple} and PromptSRC~\cite{khattak2023self} are VLM-based methods, while ViRFT~\cite{liu2025visual}, DiVE-k~\cite{kumar2026divek}, and Fine-R1~\cite{he2026finer} are MLLM-based methods. Experimental results on two zero-shot settings and the few-shot setting are reported in Table~\ref{tab:zero_shot}, Table~\ref{tab:zero_shot_mixed}, and Table~\ref{tab:few_shot}, respectively. As shown, our framework achieves the best average performance across the three settings, showing its effectiveness.

\subsection{Ablation Studies and Further Analyses}
\label{subsec:ablation}

In this section and in the supplementary, we conduct extensive ablation studies and further analyses on CUB-200~\cite{wah2011caltech}.

\begin{wraptable}{r}{0.67\linewidth}
    \centering
    \scriptsize
    \setlength{\tabcolsep}{6pt}
    \caption{Evaluation on the two key mechanisms of our framework.}
    \label{tab:ablation}
\resizebox{\linewidth}{!}{
    \begin{tabular}{lccc}
    \toprule
    Method & B & N & H \\
    \midrule
    Baseline (Qwen2.5-VL-7B~\cite{Qwen2.5-VL}) & 63.3 & 48.2 & 54.7 \\
    \midrule
    Ours (w/o MCTS-Guided Tool-Use Knowledge Distillation) & 74.3 & 68.7 & 71.4 \\
    Ours (w/o Model-Tool Co-Evolution) & 73.8 & 67.0 & 70.2 \\
    \midrule
    Ours (Full) & \textbf{84.8} & \textbf{73.0} & \textbf{78.5} \\
    \bottomrule
    \end{tabular}
}
\end{wraptable}
\textbf{Impact of the two key mechanisms.}
In our framework (\textbf{Ours (Full)}), we introduce two key mechanisms: the \emph{MCTS-Guided Tool-Use Knowledge Distillation} mechanism and the \emph{Model-Tool Co-Evolution} mechanism. To assess the contribution of each, we evaluate two variants. \textbf{(i) Ours (w/o MCTS-Guided Tool-Use Knowledge Distillation)}: we remove the training within the MCTS-Guided Tool-Use Knowledge Distillation mechanism and directly enter the Model-Tool Co-Evolution. \textbf{(ii) Ours (w/o Model-Tool Co-Evolution)}: we keep the MCTS-Guided Tool-Use Knowledge Distillation mechanism but skip the entire Model-Tool Co-Evolution. As shown in Table~\ref{tab:ablation}, both variants underperform Ours (Full), validating the efficacy of both key components in our framework.

\section{Conclusion}
\label{sec:conclusion}

In this paper, we identified that existing MLLM-based fine-grained image classification methods are \emph{not well-grounded}: their reasoning relies on vague impressions rather than the \emph{subtle yet measurable} cues that distinguish similar subcategories. To address this, we proposed \textbf{ToolFG}, the first framework that enables \emph{well-grounded} FGIC by letting the MLLM actively invoke tools to collect such cues. Extensive experiments across different experimental settings show the efficacy of our framework.

\bibliographystyle{plain}
\bibliography{egbib}

\end{document}